\documentclass{article}

\usepackage{signalpreamble}
\usepackage[preprint]{neurips_2026}
\usepackage{pgfplots}
\usepackage{tikz}
\usetikzlibrary{backgrounds , calc, shapes.geometric, arrows.meta, positioning}
\pgfplotsset{compat=1.18} 

\usepackage{listings}
\usepackage{adjustbox}
\usepackage{booktabs}
\usepackage{wrapfig}
\usepackage{caption}
\usepackage{standalone}
\usepackage{amsmath,amssymb,amsfonts,bbm}
\usepackage{placeins}

\rankcolor{first}{orange}
\rankcolor{second}{cyan}

\newcommand{\Lalgo}{\frakA}
\newcommand{\Ualgo}{\frakU}
\newcommand{\loss}{\mathsf{L}}

\newcommand{\Pdata}{\P_{\mathrm{data}}}
\newcommand{\Pret}{\P_r}
\newcommand{\Pfor}{\P_f}

\newcommand{\trainset}{{\calD_{\mathrm{train}}}}
\newcommand{\calibset}{{\calD_{\mathrm{calib}}}}

\newcommand{\forgetset}{{\calD_f}}
\newcommand{\retainset}{{\calD_r}}

\newcommand{\Xtest}{{X_{\mathrm{test}}}}
\newcommand{\Ytest}{{Y_{\mathrm{test}}}}

\newcommand{\CP}{\mathsf{C} \mathsf{P}}

\newcommand{\ncal}{n_{\mathrm{cal}}}

\newcommand{\hqalpha}{\hat{q}_{\alpha}}

\NewDocumentCommand{\Leff}{s o}{%
  \IfBooleanTF{#1}
    {\widehat{L}_{\mathrm{eff}\IfValueT{#2}{,#2}}} 
    {L_{\mathrm{eff}\IfValueT{#2}{,#2}}} 
}

\NewDocumentCommand{\Lcov}{s o}{%
  \IfBooleanTF{#1}
    {\widehat{L}_{\mathrm{cov}\IfValueT{#2}{,#2}}} 
    {L_{\mathrm{cov}\IfValueT{#2}{,#2}}} 
}

\DeclareDocumentCommand \ECF { e{_} o } {%
	C%
    \IfValueT{#1}{_{#1}}%
    \IfValueT{#2}{\parens*{#2}}%
}

\DeclareDocumentCommand \EmCF { e{_} o } {%
	M%
    \IfValueT{#1}{_{#1}}%
    \IfValueT{#2}{\parens*{#2}}%
}

\DeclareDocumentCommand \H { e{_} o } {%
	H%
    \IfValueT{#1}{_{#1}}%
    \IfValueT{#2}{\parens*{#2}}%
}

\newacronym{MU}{MU}{machine unlearning}
\newacronym{CP}{CP}{conformal prediction}
\newacronym{CMU}{CMU}{conformal unlearning}
\newacronym{DP}{DP}{differential privacy}
\newacronym{RT}{RT}{retrained model}
\newacronym{ML}{ML}{machine learning}
\newacronym{LLMs}{LLMs}{large language models}
\newacronym{cdf}{cdf}{cumulative distribution function}
\newacronym{pdf}{pdf}{probability density function}

\title{Risk-Aware Output Control Should be the Focus of Machine Unlearning}

\author{%
  Yahya Alkhatib\thanks{ya0001ib@e.ntu.edu.sg} \\
  \And
  Yuyang Xiao \\
  \And
  Wee Peng Tay \\
  \AND
  School of Electrical and Electronic Engineering \\
  Nanyang Technological University, Singapore \\
}

\begin{document}

\glsdisablehyper

\maketitle

\begin{abstract}
Well-documented failure modes of \gls{ML} systems, including bias, toxicity, hallucination, and leakage of sensitive information, has driven the rise of \gls{MU}. Most existing work, however, targets parameter-space indistinguishability, seeking unlearned models that are as close as possible to models retrained from scratch on retained data. In this paper, we argue that this objective is misaligned with practical and user-relevant concerns: \gls{MU} should prioritize statistically guaranteed output-space control rather than parameter-space similarity. We contend that output control with statistical guarantees is the operationally meaningful notion of unlearning, as it supports rigorous and robust objectives, scales to modern deployment settings, and enables interpretable guarantees. We substantiate this position with empirical evidence and with recent theoretical results that, although not originally formulated in this framing, are consistent with it.
\end{abstract}

\glsresetall{}
\section{Introduction}\label{sec:intro}

The performance and reliability of \gls{ML} systems are fundamentally determined by the quality of their training data \cite{vapnik1995nature}. Consequently, when training corpora include biased, toxic, inconsistent, sensitive, or otherwise harmful content, deployed models are likely to reproduce these pathologies in their outputs. This challenge has become more acute with vast generative models, whose scale has incentivized broad web scraping and the inadvertent collection of private and sensitive information \cite{bender2021llmsdatathirstrisksprivacy,carlini2021extractingtrainingdatallms,panda2024teachllmstophish,staab2024beyondmemorizationllms}. 

The most direct remedy is to delete the unwanted data and retrain the model from scratch. However, this is impeded by two structural constraints. First, model scale has expanded to the point that full retraining is often computationally and operationally prohibitive; architectures with trillions of parameters and hundreds of layers are increasingly common. Second, once a model is deployed, owners frequently no longer maintain complete access to the original training corpus, making faithful retraining infeasible in practice. These constraints prompted the formalization of \gls{MU} \cite{cao2015towards}: efficient procedures to satisfy deletion requests without full retraining.

Machine unlearning was initiated by \cite{cao2015towards}, who proposed removing the influence of individual training instances by transforming learning algorithms into a summation form. Since then, the literature has been strongly shaped by ideas from \gls{DP} \cite{dwork2014differential}, especially the principle that an observer should be unable to distinguish, from outputs alone, a model trained on the full dataset from one trained with a designated point -- or, more generally, a \emph{forget set} -- removed. Although this criterion originated as a privacy concept, indistinguishability has become a central objective in unlearning research \cite{guo2020indistinguishabilitycertified,mu2025certifiedrewind,chien2024certifiednoisySGD,basaran2025certifiedsourcedata,koloskova2025certifiednonconvex,izzo2021approximatedatadeletion,xu2025informationregularizationunlearning}. Two broad lines of work have emerged. The first targets \emph{exact} indistinguishability by design, typically via training procedures that support exact post hoc deletion \cite{cao2015towards,bourtoule2021SISA,brophy2021exactunlearningrandomforests}. The second adopts a pragmatic objective: approximate the model that would be obtained by full retraining on retained data empirically, and rarely with theoretical certification of approximate indistinguishability available mainly under restrictive assumptions (e.g., convex losses or linear models) \cite{guo2020indistinguishabilitycertified,izzo2021approximatedatadeletion,xu2025informationregularizationunlearning}. A concise technical overview of these approaches is provided in \cref{app:related-work}.

\textbf{This paper argues that machine unlearning should be defined and certified through rigorous statistical guarantees on the outputs of the deployed model, not through parameter-space indistinguishability from a retrained reference.} The foregoing discussion supports a central observation: \emph{an objective inherited from adversarial robustness has become the de facto standard in \gls{MU}, despite the field's broader operational aims}. A more relevant framing is whether concern should center on training provenance or on observable behavior: should one prefer a model that never trained on an individual's data yet may still reveal private information about that individual, or a model that did train on those data but is certified not to reveal such information? Our position aligns with the latter. In practice, modern deep models are used as input--output systems, while internal states -- even when inspectable -- are largely uninterpretable and provide limited evidence about user risk absent output-level analysis \cite{gilpin2018overviewofinterpretability,lipton2018mythosmodelinterpretability,carlini2019secretmemorizationineuralnets}. Consequently, output control is the more meaningful object of evaluation and assurance. We therefore identify three structural bottlenecks that motivate a shift in both objective and methodology.

\textbf{Lack of meaningful guarantees.} Certified proximity -- or even certified indistinguishability -- to a \gls{RT} does not translate into any guarantee on output behaviour: small perturbations in parameters, or even in inputs, can produce large deviations in outputs \cite{golatkar2020eternalsunshine,hsu2025unseenresidualknowledge}. The practical consequence is a self-defeating loop. Because parameter-level guarantees say so little about behaviour, most \gls{MU} methods end up benchmarking the unlearned model against an explicitly retrained one to verify that it actually behaves as intended \cite{chundawat2023bad,chundawat2023zeroshot,tarun2024UNSIR,foster2024information,graves2021amnesiac,chien2024certifiednoisySGD,golatkar2020forgettingoutsidethebox,hsu2025unseenresidualknowledge,thudi2022unrolling} -- precisely the computation that \gls{MU} was introduced to avoid. What users actually need is something different: a clear, interpretable, and quantifiable statement that speaks to their concerns, such as a probabilistic bound on how often the model emits harmful content or leaks private information after unlearning.

\textbf{Forgeability.} The work \cite{thudi2022forgeable} demonstrated that even perfect indistinguishability can be undermined through the construction of ``unlearning proofs''. A model owner could present users with a complete parameter-update history, detailing the data batches used at each training step and the corresponding parameter changes, to claim that specific data played no role in training. However, the underlying parameter trajectory could remain identical to one that did incorporate the data. This implies that a proof of ideal unlearning can be fabricated even when no actual unlearning has occurred, rendering indistinguishability an ineffective objective in practice. While one might argue that trusted third parties could authenticate such proofs to prevent forgery, this introduces additional complexities, which lead to the next limitation.

\textbf{Unreliability of the retrained gold standard.} Even when parameter-level proximity to a \gls{RT} is achieved, practical challenges persist. Generative models have been shown to reproduce copyrighted content despite never being explicitly trained on it \cite{cooper2025generativeaipolicy}, and classifiers employing \gls{CP} may still include forgotten points within their prediction sets \cite{alkhatib2026conformalunlearning, shi2025rethinking}. These examples underscore a critical limitation: \emph{indistinguishability from a \gls{RT} does not inherently guarantee the absence of undesired outputs}. When model capabilities generalize to the forget set, output-level guarantees become indispensable, as they cannot be inferred solely from parameter-level indistinguishability.

Taken together, these observations underscore the need for novel paradigms in \gls{MU} that: (1) eliminate dependence on \gls{RT}s -- or any computationally expensive baselines -- as the definitive standard; (2) are inherently resistant to falsification; and (3) provide precise, interpretable, and operationally meaningful guarantees that directly address user concerns and instill confidence in model behavior. This paradigm shift is increasingly critical as \gls{ML} systems become ubiquitous, model sizes continue to expand, and the volume of data driving these systems grows at an unprecedented pace. While this paper does not propose a fully comprehensive redefinition to resolve all these challenges, it advocates for emerging frameworks that place rigorous, distribution-free risk-control guarantees directly on a model's outputs.

A related work by \cite{cooper2025generativeaipolicy} addresses policy-makers and legislators, and moderates expectations regarding the practical capabilities of \gls{MU}. In contrast, our contribution is technically focused and directed towards researchers and practitioners, advocating for a reorientation of the field through a three-flaw structural analysis of the current paradigm and introducing concrete output-control alternatives.

The rest of this paper is organized as follows. \Cref{sec:preliminaries} reviews the necessary background. \Cref{sec:problem-statement} formalizes the problem and fixes notation. \Cref{sec:evidence-against} assembles theoretical and empirical evidence against the prevailing indistinguishability-based view. \Cref{sec:position} develops our position, introduces a distribution-free risk-control framework for output-level unlearning, and recovers conformal prediction as a canonical special case. \Cref{sec:discussion} discusses the implications of our findings and future directions. \Cref{sec:objections} addresses an alternative view to our position, and \Cref{sec:conclusion} concludes.


\section{Preliminaries}\label{sec:preliminaries}

This section presents the foundational background necessary for the paper's arguments. For clarity, we adopt a supervised learning framework, though the discussion generalizes to unsupervised and self-supervised settings with minor adjustments. We also review split \gls{CP} \cite{papadopoulos2002scp}, which serves as the primary example of an output-level guarantee when substantiating our position in \cref{sec:position}.

\paragraph{Data, models, and training.}
Let $\calX \subseteq \bbR^{d}$ and $\calY \subseteq \bbR$ denote the input and label spaces; we take scalar labels for notational simplicity, as everything that follows extends to vector-valued or structured outputs without conceptual change. A \emph{dataset} is a collection $\calD = \set{(X_i, Y_i)}_{i=1}^{N} \overset{\mathrm{i.i.d.}}{\sim} \Pdata$ drawn from a joint distribution $\Pdata$ on $\calX \times \calY$. A \emph{model} is indexed by a parameter $\theta$ in a parameter space $\Theta$ and induces a (possibly stochastic) predictor $\theta : \calX \to \calY$. With standard abuse of notation we write $\theta$ both for the parameter and the function it induces, and $\theta(X)$ for a prediction (a sample from the predictive distribution when $\theta$ is stochastic). A \emph{training algorithm} is a (possibly stochastic) map $\Lalgo : \Theta \times \bigcup_{N \ge 1} (\calX \times \calY)^{N} \times \Lambda \to \Theta$ which, given an initial model $\theta_{\mathrm{init}}$, a dataset $\calD$, and hyperparameters $\lambda \in \Lambda$, returns a trained model
\begin{align}\label{eq:family-loss-min}
    \hat{\theta} = \Lalgo(\theta_{\mathrm{init}}, \calD, \lambda) \approx \argmin_{\theta \in \Theta}  \ofrac{N} \sum_{i=1}^{N} \loss\!\left(Y_i, \theta(X_i)\right),
\end{align}
for a loss $\loss : \calY \times \calY \to \bbR_{\ge 0}$ quantifying the discrepancy between a prediction and its target. For the non-convex losses and deep architectures of interest here, $\hat{\theta}$ is only an approximate stationary point produced by an iterative optimizer such as stochastic gradient descent.

\paragraph{Split conformal prediction.}
A training algorithm returns a point estimate $\hat{\theta}$ with no accompanying statement about its predictive uncertainty, which motivates procedures that equip the model with explicit, rigorous guarantees. Among these, \emph{\gls{CP}} has emerged as particularly attractive: it is model-agnostic, distribution-free, and provides finite-sample coverage guarantees at any user-chosen level \cite{vovk2005alrw,angelopoulos2022gentle,angelopoulos2025theoretical,barber2023nonexchangeablecp}. We focus on its computationally lightweight \emph{split} variant \cite{papadopoulos2002scp}, which partitions the labelled data into a disjoint \emph{proper training set} $\trainset$ and \emph{calibration set} $\calibset$ of size $\ncal \triangleq \abs{\calibset}$, fits the model as $\hat{\theta} = \Lalgo(\theta_{\mathrm{init}}, \trainset, \lambda)$, and uses $\calibset$ -- by construction independent of $\hat{\theta}$ -- to probe its predictive uncertainty.

A \emph{nonconformity score} $s(X, Y; \hat{\theta}) \in \bbR$, with larger values indicating stronger disagreement between $Y$ and $\hat{\theta}(X)$, measures how ``atypical'' a candidate pair $(X, Y)$ is under $\hat{\theta}$. Canonical choices are $s(X, Y; \hat{\theta}) = 1 - \hat{p}_{\hat{\theta}}(Y \mid X)$ for classifiers (with $\hat{p}_{\hat{\theta}}(\cdot \mid X)$ the softmax pseudo-probability) \cite{sadinle2019least}, and $s(X, Y; \hat{\theta}) = \abs{Y - \hat{\theta}(X)}$ for regressors \cite{romano2019cqr}. Given a user-specified miscoverage level $\alpha \in (0,1)$, let $\hqalpha$ denote the $\lceil (\ncal + 1)(1-\alpha) \rceil$-th smallest of the calibration scores $\set{ s(X_i, Y_i; \hat{\theta}) \given (X_i, Y_i) \in \calibset }$ -- equivalently, their empirical $(1-\alpha)$-quantile inflated by the finite-sample factor $(\ncal + 1)/\ncal$.\footnote{When $\lceil (\ncal+1)(1-\alpha) \rceil > \ncal$, we set $\hqalpha = +\infty$ by convention.} For a fresh test input $\Xtest$, the split \gls{CP} prediction set and its marginal coverage guarantee are
\begin{align}\label{eq:cp-set-and-coverage}
    \calC(\Xtest) = \set*{ y \in \calY : s(\Xtest, y; \hat{\theta}) \le \hqalpha }, \quad \P(\Ytest \in \calC(\Xtest)) \ge 1 - \alpha,
\end{align}
where the probability is over the joint randomness of $\calibset$, the test pair $(\Xtest, \Ytest)$, and any stochasticity in $\hat{\theta}$ and $s$. The guarantee is distribution-free and holds whenever $(\Xtest, \Ytest)$ is exchangeable with $\calibset$ conditional on $\trainset$ -- in particular, whenever the labelled data are i.i.d.\ from $\Pdata$ and $\trainset$, $\calibset$ are drawn independently.

The template of \cref{eq:cp-set-and-coverage} -- a simple, interpretable, user-tunable probability that a desired event occurs -- is precisely the kind of output-level guarantee we will argue, in \cref{sec:position}, should replace parameter-level indistinguishability as the target of \gls{MU}.


\section{Machine Unlearning Problem Definition}\label{sec:problem-statement}

We now formally define the \gls{MU} problem and establish the notation used throughout the remainder of the paper. Let $\theta_o \in \Theta$ denote the \emph{original} model, obtained by applying the training algorithm to an initial model $\theta_{\mathrm{init}}$ on the full training dataset, $\theta_o = \Lalgo(\theta_{\mathrm{init}}, \trainset, \lambda)$, with $\lambda$ a collection of training hyperparameters. We assume that the training set admits a disjoint partition $\trainset = \retainset \sqcup \forgetset$, into a \emph{retain set} $\retainset$, whose influence on the model is to be preserved, and a \emph{forget set} $\forgetset$, whose influence is to be removed. For later use, we denote by $\Pret$ and $\Pfor$ the marginal distributions associated with the retain and forget regions of the input-label space, respectively. When the forget set corresponds to an identifiable subpopulation---for example, a particular user, class, or topic---$\Pfor$ is naturally interpreted as the conditional distribution restricted to that subpopulation.

A \emph{\gls{MU} algorithm}, denoted $\Ualgo$, is a (possibly stochastic) map that takes as input the original model, the retain and forget sets, and a collection of algorithm-specific hyperparameters $\lambda_u$, and produces an \emph{unlearned model} $\theta_u = \Ualgo(\theta_o, \retainset, \forgetset, \lambda_u) \in \Theta$, whose behavior is required to reflect the influence of $\retainset$ while suppressing that of $\forgetset$, in a sense made precise by the chosen unlearning criterion.

The canonical baseline is the \emph{retraining algorithm} $\Ualgo_{\mathrm{RT}}$, which reinitializes training from $\theta_{\mathrm{init}}$ using only the retain set, entirely discarding both the original model and the forgotten data, yielding the \emph{\gls{RT}} $\theta_{\mathrm{RT}} \triangleq \Ualgo_{\mathrm{RT}}(\theta_o, \retainset, \phi, \lambda) = \Lalgo(\theta_{\mathrm{init}}, \retainset, \lambda)$. By construction, $\Ualgo_{\mathrm{RT}}$ is independent of both $\theta_o$ and $\forgetset$. The \gls{RT} $\theta_{\mathrm{RT}}$ has consequently been adopted as the de facto gold standard for \gls{MU}: a model trained exclusively on $\retainset$ is, by definition, free of any influence from $\forgetset$, and is therefore regarded as the ideal target that any unlearning procedure should seek to emulate.

A dominant formalization of this intuition posits that an observer with access to the outputs of an unlearned model $\theta_u$ should not be able to extract more information about the forgotten data than an observer of the outputs of the \gls{RT} $\theta_{\mathrm{RT}}$, which, by construction, has never been trained on those points \cite{guo2020indistinguishabilitycertified,ginart2019exactkmeansunlearning,cao2015towards,shokri2017membership}. In practice, this principle is typically instantiated as a differential-privacy-inspired indistinguishability constraint on the distribution of the returned model.

\begin{Definition}[$(\epsilon,\delta)$-indistinguishability \cite{guo2020indistinguishabilitycertified}]\label{def:indistinguishability}
An unlearning algorithm $\Ualgo$ is said to be $(\epsilon,\delta)$-indistinguishable from the retraining algorithm $\Ualgo_{\mathrm{RT}}$ if, for every measurable set $\calT \subseteq \Theta$,
\begin{align}\label{eq:indistinguishability}
    \P(\Ualgo(\theta_o, \retainset, \forgetset, \lambda_u) \in \calT) &\le e^{\epsilon}\P(\Ualgo_{\mathrm{RT}}(\theta_o, \retainset, \phi, \lambda) \in \calT) + \delta, \\
    \P(\Ualgo_{\mathrm{RT}}(\theta_o, \retainset, \phi, \lambda) \in \calT) &\le e^{\epsilon}\P(\Ualgo(\theta_o, \retainset, \forgetset, \lambda_u) \in \calT) + \delta,
\end{align}
where the probabilities are taken over the (possibly stochastic) randomness of the two algorithms.
\end{Definition}

Several variants of this idea have been proposed. For example, \cite{xu2025informationregularizationunlearning} adopt a ``flipped'' formulation in which the information carried by the inputs of the unlearned model about its outputs is required to be no larger than the corresponding quantity for the \gls{RT}. The \gls{RT} thus plays the role of an unquestioned \emph{golden baseline}, and the unlearning problem collapses into an exercise in approximating it. We argue in the remainder of the paper that this paradigm is both limiting and, in crucial respects, ill-posed.


\section{Evidence Against the Current Stance}\label{sec:evidence-against}

Having set out the prevailing indistinguishability-based programme in \cref{sec:problem-statement}, we now turn to a more fundamental question: is the programme sound \emph{in principle}? We argue that it is not. This section identifies three distinct structural deficiencies in the current paradigm: (i) proximity in parameter space does not imply proximity in model behavior; (ii) indistinguishability certificates are, in general, constructively defeatable; and (iii) the \gls{RT} -- the very gold standard against which unlearning is measured -- can itself fail to guarantee the desired model behavior of the unlearning objective. These deficiencies are mutually reinforcing: resolving any two leaves the third fully intact, and the third alone is sufficient to necessitate a shift towards an output-centric formulation of \gls{MU}.

\paragraph{Flaw I: the parameter--output gap.} 
The indistinguishability objective -- i.e., driving $\theta_u$ as close as possible to $\theta_{\mathrm{RT}}$ in parameter space, whether approximately or exactly -- implicitly assumes that parameter proximity entails behavioral proximity. This assumption is generally false, and the failure is not an artifact of pathological non-continuous parameterizations: it occurs for smooth, fully continuous model classes. The simplest illustration is the elementary linear model $f_\theta(x) = a\cdot b\cdot x$ with $\theta = (a,b)$. Perturbing only $b$ by $\delta$ yields $\|\theta - \tilde\theta\|_2 = |\delta|$ but produces an absolute output change $|f_\theta(x) - f_{\tilde\theta}(x)| = |a\,\delta\,x|$, which can be made arbitrarily large for fixed $\delta$ by an appropriate choice of $a$. This phenomenon is structural: Theorem~1 of \cite{bernstein2020parameterdistanceandstabilityoflearningMLP} establishes that for an $L$-layer multilayer perceptron the relative functional change satisfies
\begin{equation}
\frac{\|f_{\tilde\theta}(x) - f_\theta(x)\|_2}{\|f_\theta(x)\|_2} \le \parens*{\frac{\gamma}{\nu}\kappa^2}^L \brk*{\prod_{k=1}^{L}\parens*{1 + \frac{\|\Delta W_k\|_F}{\|W_k\|_F}} - 1},
\end{equation}
where $\nu, \gamma$ govern the bi-Lipschitz behavior of the nonlinearity and $\kappa$ bounds the condition number of the weight matrices $W_k$ and their perturbations $\Delta W_k$; the toy example above saturates the bound (\cite{bernstein2020parameterdistanceandstabilityoflearningMLP}, Eq.~3, with $\Delta a = 0$). Crucially, \cite{bernstein2020parameterdistanceandstabilityoflearningMLP} also \emph{empirically} verify that the actual relative change of a $16$-layer MLP grows quasi-exponentially in the perturbation size at every training checkpoint they probe (their Figure~1), establishing that the phenomenon is not a vacuous worst-case artifact but observed behavior in practical deep networks. Consequently, an arbitrarily small $\ell_2$ displacement in parameter space can induce arbitrarily large changes in the function the model implements. The converse failure -- large $\ell_2$ displacement that nevertheless leaves behavior unchanged -- is equally well documented. In the unlearning context specifically, \cite{golatkar2020forgettingoutsidethebox} argue that overparameterized networks possess a large \emph{null-space} of weights producing identical activations on all test samples; consequently, the weight-space distance $\|\theta_u - \theta_{\mathrm{RT}}\|_2$ can be large while the input--output behaviors are indistinguishable, and they accordingly propose evaluating forgetting in \emph{activation space} rather than parameter space. Together, these two observations show that $\ell_2$ proximity in parameter space is neither necessary nor sufficient for behavioral equivalence -- even for the smooth, well-behaved architectures used in practice.

\begin{figure}[!tb]
\centering
\begin{subfigure}[t]{0.20\textwidth}
\centering
\includegraphics[width=\linewidth]{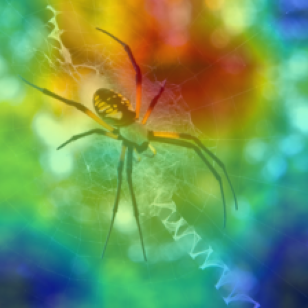}
\caption{Gradient-clipping overlay}
\label{fig:PABI_gradcam}
\end{subfigure}
\begin{subfigure}[t]{0.20\textwidth}
\centering
\includegraphics[width=\linewidth]{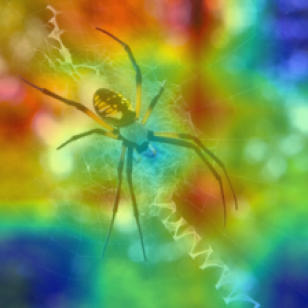}
\caption{RT overlay}
\label{fig:RT_gradcam}
\end{subfigure}
\begin{subfigure}[t]{0.20\textwidth}
\centering
\includegraphics[width=\linewidth]{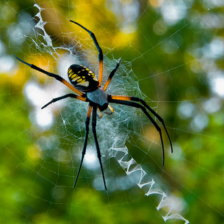}
\caption{Original image}
\label{fig:orig_image}
\end{subfigure}
\caption{Grad-CAM overlay regions on an image after forgetting. Certified unlearning gradient-clipping generates a saliency map significantly different from that of RT.}
\label{fig:saliency-deviation}
\end{figure}

The implication for unlearning is immediate: elevating $\theta_{\mathrm{RT}}$ to a golden standard and approximating it, however tightly, licenses no claim about the unlearned model's actual \emph{behavior}. \Cref{fig:saliency-deviation} demonstrates that the saliency map produced, on a forgotten image, by gradient-clipping \cite{koloskova2025certifiednonconvex} -- a certified unlearning method in the sense of \cref{def:indistinguishability} trained on $\retainset$ for the same number of epochs as a \gls{RT} -- visibly departs from the saliency map produced by the \gls{RT}, despite gradient-clipping's explicit guarantees of proximity to that model. \Cref{tab:perturbation-accuracy} sharpens the point: two otherwise identical models that differ only by a small weight perturbation exhibit non-trivial accuracy drops and prediction disagreement on the same test samples (see \cref{app:first-layer-signflip} for details and \cref{app:extra-result-gaussian-noise} for further experiments). Small parameter perturbations are simply not behaviorally small, and so approximate indistinguishability cannot be expected to translate into approximate behavioral equivalence. Since parameter-level proximity cannot be trusted to imply behavioral proximity, practitioners routinely fall back on comparisons against the \gls{RT} itself to verify that the unlearned model behaves as intended -- precisely the computation that \gls{MU} was introduced to avoid.

\begin{table}[!htbp]
\centering
\caption{Accuracy drop and top-1 prediction disagreement ($\%$) for five pretrained models subjected to random first-layer weight sign-flipping ($\norm{\theta_o - \theta_{\mathrm{perturb}}}/\norm{\theta_o} \le 2\%$).}
\label{tab:perturbation-accuracy}
\footnotesize
\begin{tabular}{lccccc}
\toprule
& ResNet-50 & RepVGG-A2 & ViT-B/16 & ConvNeXt-T & EffNet-B0 \\
\midrule
Accuracy Drop (\%)    & 19.52 & 71.40 & 11.90 & 81.86 & 15.34  \\
Disagreement (\%) & 34.08 & 94.22 & 23.58 & 99.14 & 30.36 \\
\bottomrule
\end{tabular}
\end{table}

\paragraph{Flaw II: indistinguishability is forgeable.}
Suppose, for the sake of argument, that the parameter--output gap could be bridged and that a method delivered \emph{perfect} indistinguishability. Even then, the objective would be practically empty: a model owner could provide the user with a convincing proof of unlearning without performing any unlearning at all. This is the content of Theorem~3 of \cite{thudi2022forgeable}, stated for minibatch stochastic gradient descent.

\begin{Theorem}[Existence of a forging adversary \cite{thudi2022forgeable}]\label{th:any-dataset-is-forgeable}
Let an adversary have freedom over the choice of minibatches $b$ and the ability to sample indefinitely from a dataset $\calD$ under a measure that places zero mass on any individual point. Then for every $\varepsilon > 0$, for almost every dataset $\calD$ there exists a finite dataset $\tilde{\calD}$ whose training trajectory can be forged by a finite dataset $\calD'$ disjoint from $\tilde{\calD}$.
\end{Theorem}

Forging here means that the training trajectories (parameter updates) of $\tilde{\calD}$ and $\calD'$ are identical. Concretely, a model owner can hand the user a complete parameter-update history -- the minibatch and the parameter change at every step -- that ``proves'' the user's data were never used, while the underlying trajectory is identical to one that \emph{did} use the data. Both approximate \emph{and} exact unlearning (i.e., a ($\epsilon=0$,$\delta=0$)-indistinguishable model), when judged through the indistinguishability lens, are thus falsifiable. This is not a technical defect of particular algorithms but a fundamental limitation of the notion itself: by reducing unlearning to a statement about whether particular data were used during training, indistinguishability invites precisely the kind of proof it was meant to preclude.

\paragraph{Can flaws I and II be patched?}
One might hope to circumvent the first two flaws by (a) improving algorithms to achieve exact rather than approximate indistinguishability, and (b) delegating the certification of unlearning to a trusted third party capable of detecting forgeries. Both are reasonable engineering responses. However, the third flaw is not amenable to such repairs, for it strikes at the gold standard itself.

\begin{figure}[!htbp]
  \centering
  \begin{subfigure}[b]{0.34\linewidth}
    \centering
    \includegraphics[width=1.0\linewidth]{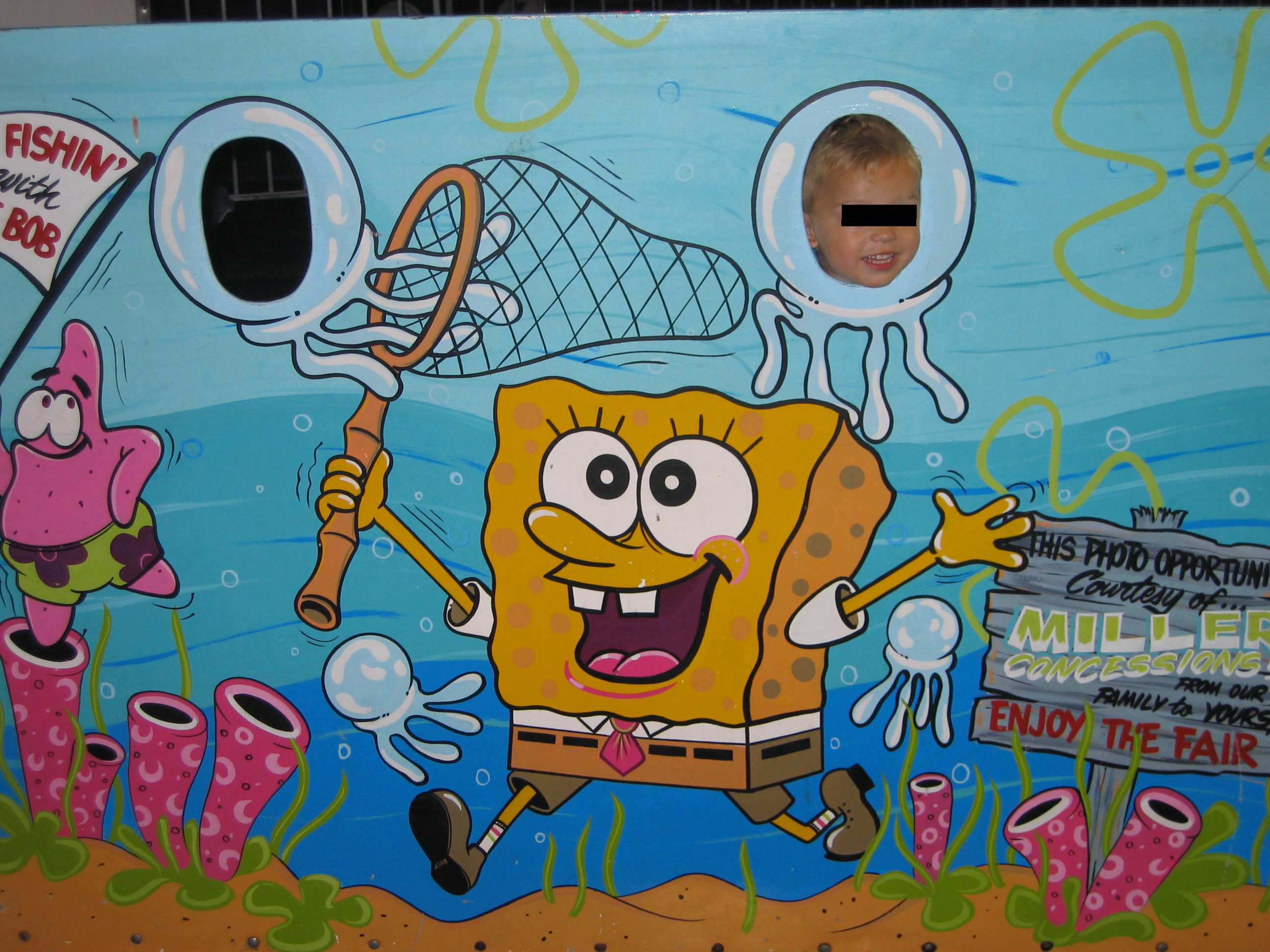}
    \caption{SpongeBob Squarepants dataset instance (censor bar). \phantom{, full body, cheerful, white background}}
    \label{fig:spongebob-a}
  \end{subfigure}
  \hspace{0.5cm}
  \begin{subfigure}[b]{0.27\linewidth}
    \centering
    \includegraphics[width=0.95\linewidth]{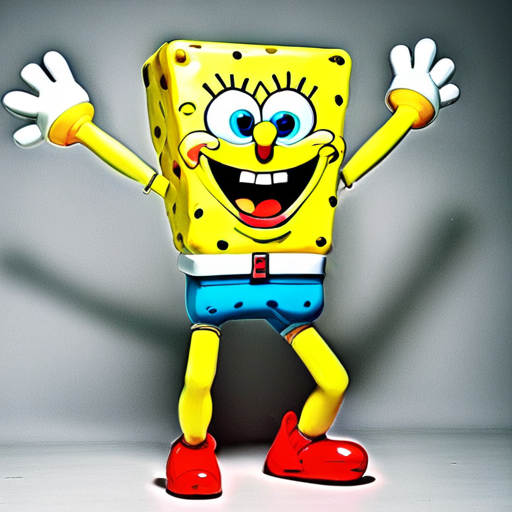}
    \caption{Prompt: \textit{``SpongeBob SquarePants, full body, cheerful, white background''}.}
    \label{fig:spongebob-b}
  \end{subfigure}
  \hspace{0.5cm}
  \begin{subfigure}[b]{0.27\linewidth}
    \centering
    \includegraphics[width=0.95\linewidth]{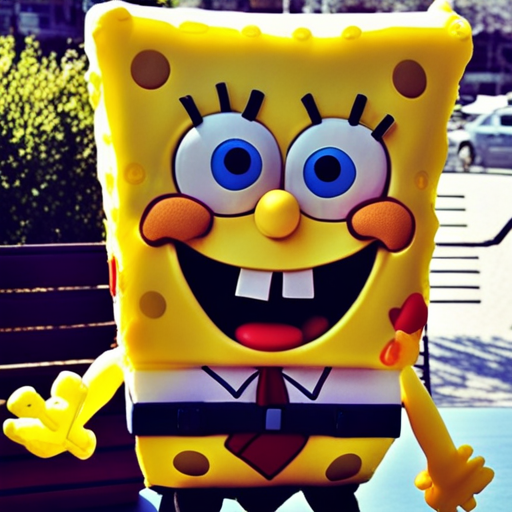}
    \caption{Prompt: \textit{``SpongeBob SquarePants''}.\phantom{, full body, cheerful, white background} \phantom{----------------------}}
    \label{fig:spongebob-c}
  \end{subfigure}
  \caption{SpongeBob Squarepants imagery generated by CommonCanvas-XL-C.}
  \label{fig:generative-copyright-sponge}
\end{figure}

\paragraph{Flaw III: the retrained model is not a reliable standard.}
The indistinguishability paradigm treats the \gls{RT} $\theta_{\mathrm{RT}}$ as an idealized target: a model that, by construction, has never seen the forgotten data and therefore ``should not'' display any undesired behavior attributable to it. Recent evidence contradicts this premise. Generative models trained without specific data samples routinely generalize into the very regions of input space those samples occupied, producing outputs that infer or reproduce information the model was never explicitly trained on \cite{gandikota2023generativeunlearning,cooper2025generativeaipolicy}. We reproduce similar examples to those in \cite{cooper2025generativeaipolicy}. \Cref{fig:generative-copyright-sponge} shows a generative image model producing copyright-infringing imagery of SpongeBob absent from the model's training data, which are carefully collected as Common-Licensed \cite{gokaslan2024commoncanvas} (more in \cref{app:copy-right-extra-results}). A similar effect has been reported for \gls{LLMs} that unlearned Harry Potter novels but could still infer a lot of information about those stories from open-source sources such as Wikipedia articles \cite{eldan2023whos} (cf. \cref{tab:hp-leak} in \cref{app:rep-hp}). A complementary manifestation of the same phenomenon appears in classification: \gls{CP} sets produced by a \gls{RT} continue, with non-trivial probability, to cover the labels of the very points that were excluded from training \cite{alkhatib2026conformalunlearning,shi2025rethinking} (see \cref{tab:rt-forget-coverage}).

The consequence for the indistinguishability framework is stark. If, without ever having been trained on the offending data, the \gls{RT} itself can produce copyright-infringing, private, or otherwise harmful outputs or can correctly cover the forgotten data, e.g. in the \gls{CP} sense, then there is simply \emph{nothing to unlearn} in the indistinguishability sense: the dataset is already ``clean'', yet the model still misbehaves.
Removing further samples would only impair the model's legitimate capabilities without addressing the actual problem. Yet a user whose copyright has been infringed or who has received harmful generated content is still entitled to demand a corrective operation. Indistinguishability-based unlearning has no answer here. A notion that cannot accommodate this scenario is a notion that has misidentified what unlearning is for.

\paragraph{From flaws to a constructive pivot.}
What concerns users, model owners, and practitioners is not whose data appeared in the training set but what the deployed model actually \emph{produces}. Once unwanted data have been deleted from storage, the questions that remain -- does this model emit my private information? does it generate copyright-infringing content? does it output biased or toxic responses? -- are questions about the model's outputs, not its parameters. This suggests a principled split of the research agenda: one line concerned with \emph{output control}, delivering rigorous statistical guarantees on what the model generates; and a complementary line concerned with \emph{parameter-level robustness}, ensuring that adversaries cannot reconstruct training samples from the model's weights. The first of these lines -- which side-steps all three flaws identified above -- is precisely the one we argue has been neglected, and is the subject of the next section.


\section{Our Position: Output Control Yields Operationally Meaningful Machine Unlearning}\label{sec:position}

\paragraph{What we advocate.}
Building on the critique in \cref{sec:evidence-against}, we advocate a shift in the \gls{MU} agenda toward objectives defined at the level of observable model behavior. We argue for prioritizing definitions of unlearning that are operationally meaningful, technically tractable, and directly aligned with deployment outcomes: what the system \emph{outputs}, rather than how closely its parameters match an idealized retrained reference. Under this view, the central requirement is explicit statistical control of undesirable outputs, independent of weight-space proximity. For example, can a model owner certify that a deployed pipeline emits user-specific private information with probability at most $1\%$, or that harmful, discriminatory, or copyright-infringing content arises with probability at most $5\%$? Uncertainty quantification offers a template. Established frameworks -- false discovery rate control \cite{benjamini1995controllingfdr, barber2015controllingfdrknockoffs,ramdas2017controlfdrdecayingmemory}, risk control \cite{bates2021distributionriskcontrol, angelopoulos2025LTT}, and conformal coverage guarantees \cite{vovk2005alrw,shafer2008conformal,angelopoulos2022gentle,lei2014distribution} -- yield rigorous, user-tunable probabilistic statements about model outputs. Machine unlearning requires an analogous framework with an inverted objective: for example, beyond covering true labels on data it \emph{should} retain, the model should also exhibit elevated miscoverage (or, more generally, controlled failure) on data it should forget, while preserving its guarantees on the retained distribution.

\paragraph{A general risk-control framework for unlearning.}
The harms that motivate unlearning in practice -- toxicity, bias, hallucination, privacy leakage, copyright infringement -- span output modalities well beyond any single metric. A principled generalization replaces any specific output event with a user-chosen risk functional. The learn-then-test (LTT) framework of \cite{angelopoulos2025LTT} achieves model-agnostic, distribution-free control of arbitrary risks, supplying precisely the theoretical tool we use.

Let $g : \calY \times \calY \to \bbR$ be a \emph{grading function} that scores a prediction $\theta(X)$ against the ground truth $Y$ along the axis of interest, with larger values encoding a more desirable output. For a threshold $\lambda \in \bbR$, the associated \emph{admission function} is
\begin{align}\label{eq:admission-function}
    a(\theta(X), Y) \triangleq \indicate*{ g(\theta(X), Y) \ge \lambda } \in \{0, 1\},
\end{align}
taking the value $1$ when the output is admissible and $0$ otherwise.

\begin{Definition}[$(\alpha, \beta)$-output-control unlearning]\label{def:output-controlMU}
For $0 \le \alpha \le \beta \le 1$, an unlearning algorithm $\Ualgo$ is said to be $(\alpha, \beta)$-\emph{output-control unlearning} for an admission function $a$ if the unlearned model $\theta_u$ satisfies
\begin{align}
    \Pret\bigl(a(\theta_u(X), Y) = 1\bigr) & \ge 1 - \alpha, \label{eq:output-control-retain} \\
    \Pfor\bigl(a(\theta_u(X), Y) = 0\bigr) & \ge \beta. \label{eq:output-control-forget}
\end{align}
\end{Definition}

What is instructive about \cref{def:output-controlMU} is what it does \emph{not} say. It imposes no proximity requirement between $\theta_u$ and any retrained reference: the unlearned model is free to differ substantially from $\theta_{\mathrm{RT}}$, provided \cref{eq:output-control-retain,eq:output-control-forget} hold. The unlearning problem becomes one of driving the forget-set admission rate below $1 - \beta$ while preserving the retain-set admission rate above $1 - \alpha$ -- two statements placed directly on model outputs rather than on weights. The choice of $g$ (and hence $a$) is application-specific and can encode more than one risk: if $g$ is a human- or AI-based grader scoring how biased, toxic, hallucinated, copyright-infringing, or privacy-leaking a prediction is, \cref{def:output-controlMU} delivers a direct guarantee against the very behaviors that motivate the unlearning request. Frameworks such as LTT \cite{angelopoulos2025LTT} provide a concrete route to certifying such guarantees from a finite calibration set, with $\lambda$ chosen to meet the target levels $\alpha$ and $\beta$. Conformal prediction yields a natural special case, which we develop next.

\begin{Remark}[Output control is not just output suppression.]
Output control, in the sense of \cref{def:output-controlMU}, is \emph{not} synonymous with inference-time output suppression. The latter treats the model as fixed and confines intervention to filters, refusals, or re-ranking of its generations, narrowing both the scientific question and the class of admissible remedies. Output control instead specifies a \emph{behavioral property} -- a user-chosen risk functional with distribution-free guarantees -- while remaining agnostic to the mechanism: parameter updates, architectural changes, auxiliary modules, inference-time interventions, or any combination thereof. Inference-time unlearning is thus a \emph{special case} of output control, and remains a valid and active line of research (cf.\ \cite{chowdhury2026inferencetimeunlearningusingconformal}).
\end{Remark}

\paragraph{Conformal unlearning as a special case.}
The conformal-prediction-based formulation of \cite{alkhatib2026conformalunlearning,shi2025rethinking} arises as a canonical instance of \cref{def:output-controlMU}. Our point of departure is the finding of \cite{shi2025rethinking} that an unlearned model---including one obtained via full retraining---may still assign \gls{CP} sets that contain the true labels of forgotten examples, a behavior they describe as ``fake unlearning.'' This motivates specializing \cref{def:output-controlMU} to the coverage event. We denote by $\CP(\theta, s)$ the split \gls{CP} procedure from \cref{sec:preliminaries} with base model $\theta$ and nonconformity score $s$. Setting $g(\theta(X), Y) = -s(X,Y;\theta)$---the conformity scoring of the deviation between the prediction and the true label---and $\lambda = - \hqalpha$, the admission function reduces to $a(\theta(X), Y) = \indicate{Y \in \calC_\theta(X)}$, and \cref{def:output-controlMU} specializes to the requirement that the unlearned model satisfy
\begin{align}
    \Pret(Y \in \calC_{\theta_u}(X)) & \ge 1 - \alpha, \label{eq:conformal-coverage-retain} \\
    \Pfor(Y \notin \calC_{\theta_u}(X)) & \ge \beta, \label{eq:conformal-coverage-forget}
\end{align}
recovering exactly the $(\alpha, \beta)$-\emph{\gls{CMU}} of \cite{alkhatib2026conformalunlearning}. If $\Pfor$ were known exactly, \cref{eq:conformal-coverage-forget} could be satisfied trivially by setting $\calC_{\theta_u}(X)=\emptyset$ on every $X$ from its marginal -- but this is not operationally meaningful. In practice the model owner has only a finite sample from $\Pfor$. Moreover, mapping all forget inputs to a fixed output would render them readily identifiable to an external observer, undermining the very privacy or security motivations for unlearning. Output-control objectives must therefore be met not by ad hoc ``if-else'' rules but by statistically principled methods that remain practical under partial knowledge of the forget distribution. In that case, achieving the joint requirements of \cref{eq:conformal-coverage-retain,eq:conformal-coverage-forget} is non-trivial: \cref{tab:rt-forget-coverage} shows that despite a significant accuracy drop on the forget set, the \gls{RT} itself exhibits essentially no change in coverage over it. We stress that conformal unlearning is \emph{one} instantiation of \cref{def:output-controlMU}, not its endpoint: many other admissible specializations are possible, depending on the application and the risk one wishes to control.

\setlength{\intextsep}{0pt}
\begin{wraptable}[8]{r}{0.40\textwidth}
\centering
\caption{Coverage and accuracy scores between an original model and RT on ImageNet at level $\alpha=0.05$.}
\label{tab:rt-forget-coverage}
\tiny
\begin{tabular}{lcc}
\toprule
& Accuracy on $\forgetset$ & Miscoverage on $\forgetset$ \\
\midrule
Original (\%) & 84.06 $\pm$ 0.00 & 0.06 $\pm$ 0.00\\
RT (\%)       & 33.07 $\pm$ 4.00 & 0.08 $\pm$ 0.02 \\
\bottomrule
\end{tabular}
\end{wraptable}

Conformal unlearning extends naturally to membership inference attacks (MIAs) \cite{shokri2017membership}: by treating the membership of a forgotten point as a binary label, one defines a membership risk and controls it at a user-specified level. \cite{shi2025rethinking} introduce the MIA conformal ratio (MIACR) for exactly this purpose, measuring the fraction of forgotten points whose membership is miscovered by a binary \gls{CP} set. Methods that control this risk thus address an important aspect of the \gls{MU} literature directly.

Taken together, \cref{def:output-controlMU} and its conformal instantiation illustrate the right target for \gls{MU} research: definitions whose parameters are \emph{numbers a user can read and act upon} -- a retain admission rate of at least $1 - \alpha$ and a forget admission rate of at most $1 - \beta$ -- rather than a probabilistic distance from a reference model that, as \cref{sec:evidence-against} has shown, is neither observable, forgery-proof, nor behaviorally reliable. Our position is not a verdict on particular algorithms but a call to redefine unlearning, and an invitation to develop methods that achieve, extend, or approximate the objectives of \cref{def:output-controlMU}.


\section{The Picture that Emerges}\label{sec:discussion}

The evidence in \cref{sec:evidence-against} and the alternatives in \cref{sec:position} converge on a single conclusion: the community must redirect toward objectives whose guarantees are actionable and interpretable for model owners and users alike. Output-level statistical guarantees will unlock tighter bounds, sharper objectives, and evaluation metrics that measure what actually matters. This shift preserves the essence of \gls{MU} as a data-centric model-cleaning scheme: identifying data to retain and forget remains indispensable; what changes is what we ask of the algorithm once that identification has been made.

Definitions of the form in \cref{def:output-controlMU} side-step each of the three flaws by construction:
\begin{itemize}
    \item \emph{Flaw I} (parameter--output mismatch) is avoided because guarantees are placed on the model's predictive behavior -- coverage, admission rate, risk -- not on a parameter-space distance.
    \item \emph{Flaw II} (forgeability) is avoided because the model owner supplies a statistical claim about outputs, auditable by any party with access to a labelled test stream from the retain and forget distributions; there is no training-trajectory proof to forge.
    \item \emph{Flaw III} (unreliability of the retrained gold standard) is avoided because no retrained reference is invoked: if the unlearned model admits forbidden outputs too often, it fails the definition, regardless of whether a \gls{RT} would have done any better. 
\end{itemize}  
These are structural properties of the definitions themselves.

We do \emph{not} claim that parameter-level analyses or adversarial-robustness research are misdirected. We endorse a natural split: output control, on one side, guarantees behavioral properties of the deployed system; parameter-level robustness, on the other, ensures adversaries cannot reconstruct or infer training samples from the weights. The two are necessary and complementary. What we \emph{do} claim is that, under the present orthodoxy, the first has been conflated with -- and ultimately suffocated by -- the second, and that disentangling them is essential for a \gls{MU} paradigm capable of matching the scale and stakes of modern deployed models.

\section{Objections to Output Control}\label{sec:objections}
Three objections to the output-control framework warrant a direct response. \textit{(1) Regulatory provenance.} Data-protection rights (e.g., under GDPR) are defined primarily in terms of data processing rather than model outputs; therefore, output-level guarantees alone may not satisfy legal standards that require non-use of personal data. Our position, however, is that the motivation for \gls{MU} extends beyond GDPR compliance. As argued in \cref{sec:discussion}, parameter-level robustness should be pursued as a complementary line of work to address provenance requirements directly. \textit{(2) Risk-functional specification.} Output control requires specifying a grading function $g$ \emph{a priori}, whereas indistinguishability is often presented as risk-agnostic. In light of Flaws~I and~III, this apparent universality is misleading: indistinguishability obscures the operative threat model rather than removing it, while an explicit $g$ makes the target harm transparent and allows guarantees to be tailored to the application. \textit{(3) Calibration shift.} Output-level guarantees commonly rely on exchangeability between calibration and deployment data, an assumption that can fail under adversarial probing of the forget set. This limitation is substantive but methodologically manageable through techniques for shift-robust \cite{podkopaev2021labelshiftcp,tibshirani2019covariateshift}, non-exchangeable \gls{CP} \cite{barber2023nonexchangeablecp}, and conformal risk control under distribution shift \cite{farinhas2024nonexchangeableconformalriskcontrol}.


\section{Conclusion}\label{sec:conclusion}
We have argued that the field's pursuit of indistinguishability from a \gls{RT} is a misdirection of effort, and that \gls{MU} should shift to \emph{output control} with clear, actionable, and rigorous statistical guarantees. This position rests on three structural limitations of the prevailing paradigm -- the parameter--output mismatch, the forgeability of indistinguishability certificates, and the unreliability of the \gls{RT} as a gold standard -- none patchable by engineering improvements alone. Beyond critique, the output-control framework of \cref{def:output-controlMU}, with conformal prediction as a canonical special case, shows that output-level guarantees are not only desirable but achievable. We hope this paper motivates the community to rethink its definitions and objectives through the lens of statistical guarantees placed where they matter most -- on what the model actually produces.


\bibliographystyle{plain}
\bibliography{papers,books}


\newpage

\appendix

\section{Related Work}\label[Appendix]{app:related-work}

This appendix gives a short technical survey of the machine unlearning literature. As outlined in the main text, existing methods fall into two broad categories: \emph{exact} unlearning and \emph{approximate} unlearning.

\paragraph{Exact unlearning.}
Methods in this category produce a model identical, or statistically indistinguishable, to a \gls{RT} by construction, but more efficiently than naive retraining. The most prominent example is SISA \cite{bourtoule2021SISA}, which shards the training data into disjoint subsets, trains a separate submodel on each shard, and aggregates the submodels into a final predictor; when an unlearning request arrives, only the submodels associated with shards containing the forget points are retrained from scratch, and the aggregation is then redone. \cite{brophy2021exactunlearningrandomforests} apply a similar idea to random forests. A second line of exact methods stores information about the training trajectory and rolls back the exact updates caused by the forget set; this idea was introduced by \cite{cao2015towards} for statistical query learning and adapted to exact $k$-means unlearning by \cite{ginart2019exactkmeansunlearning}. While elegant, exact methods typically require a particular training procedure and therefore do not apply to pretrained foundation models.

\paragraph{Approximate unlearning.}
Methods in this category aim to produce an unlearned model that approximates the retrained one. The certified-indistinguishability style of \cref{def:indistinguishability} gives such methods explicit room to deviate from the retrained model, parameterized by $(\epsilon, \delta)$. Most algorithms proposed under this umbrella do not actually establish such guarantees and instead rely on heuristic proxies to induce unlearning. These include influence functions \cite{guo2020indistinguishabilitycertified}, synaptic dampening of weights identified via Fisher information \cite{foster2024SSD}, knowledge distillation from a frozen teacher \cite{chundawat2023bad}, KL-divergence maximization on the forget set \cite{kurmanji2023SCRUB}, loss-maximising noise injection \cite{tarun2024UNSIR}, residual-knowledge-regularized gradient ascent \cite{hsu2025unseenresidualknowledge}, mislabelling of the forget set followed by fine-tuning \cite{graves2021amnesiac}, and information-theoretic regularization that minimizes the mutual information between the forget set and the model's predictions \cite{foster2024information,xu2025informationregularizationunlearning}, among others. Several of these strategies have been extended to \gls{LLMs} through, for instance, neuron pruning \cite{liu2025modalityawareneuronpruningMLLMsunlearning}, gradient ascent \cite{yao2024unlearningLLMsGA}, module-masked gradient ascent \cite{bao2025unlearningLLMsmodulemaskingGA}, low-rank-adapter-based gradient ascent \cite{cha2025unlearningLLMsLoRA}, weighted-loss training \cite{li2024loreunweightedlossunlearningLLMs}, negative-preference optimization \cite{zhang2024unlearningLLMsNPO}, among others.

A subset of approximate methods does establish formal guarantees of proximity in the sense of \cref{def:indistinguishability}. \cite{koloskova2025certifiednonconvex} achieve certified unlearning for non-convex losses through gradient and parameter clipping followed by fine-tuning on the retained data. For strongly convex losses, certified methods have been developed by \cite{guo2020indistinguishabilitycertified,chien2024certifiednoisySGD,basaran2025certifiedsourcedata}. Other approaches, such as \cite{mu2025certifiedrewind}, certify unlearning by repeating a controlled number of additional training steps. All of these methods, implicitly or explicitly, treat indistinguishability from a retrained reference as the gold objective.

\paragraph{Departures from indistinguishability.}
A small but growing body of work questions whether indistinguishability should remain the dominant objective. \cite{shi2025rethinking} identify the phenomenon of \emph{fake unlearning}, in which even a retrained model can still cover forget points in its conformal prediction sets despite never having been trained on them. Building on this observation, \cite{alkhatib2026conformalunlearning} formalize a \emph{conformal unlearning} scheme in which unlearning is defined directly through coverage and miscoverage requirements on the model's prediction sets (cf.\ \cref{sec:position}); they stop short, however, of providing certified procedures to attain those requirements. The position we advance in this paper is in the same spirit but broader in scope: it argues that the field as a whole should take such output-level reformulations as the starting point of unlearning research, rather than as exceptions to the indistinguishability orthodoxy.

\section{Experimental Details}\label[Appendix]{app:reproducibility}

This appendix collects the implementation details necessary to reproduce the experiments reported in the main text.

\subsection{First-layer sign-flip perturbations}\label[Appendix]{app:first-layer-signflip}

Each model is evaluated on a fixed test split of $5{,}000$ ImageNet validation images.
We consider five pretrained vision models: ResNet-50, RepVGG-A2, ViT-B/16, ConvNeXt-T, and EffNet-B0.
For each model, we perturb only the parameters in the first layer and keep all remaining parameters unchanged.
Specifically, for each first-layer parameter $\theta_i$, we independently sample
$z_i \sim \mathrm{Bernoulli}(p)$ and construct the perturbed parameter
\[
\theta_i^{\mathrm{perturb}} =
\begin{cases}
-\theta_i, & z_i = 1,\\
\theta_i, & z_i = 0.
\end{cases}
\]
Thus, each selected first-layer parameter has probability $p$ of having its sign flipped.
We sweep $p \in \{0.005, 0.01, 0.02, 0.05, 0.075, 0.1, 0.15, 0.2, 0.3, 0.5\}$ and compute the global relative parameter distance
\[
d_{\mathrm{rel}} =
\frac{\|\theta_o - \theta_{\mathrm{perturb}}\|_2}{\|\theta_o\|_2}.
\]
For each model, we report the run with the largest $d_{\mathrm{rel}}$ satisfying $d_{\mathrm{rel}} \leq 2\%$.
The perturbed first-layer modules are \texttt{conv1} for ResNet-50, \texttt{stem.conv\_kxk.conv} for RepVGG-A2, \texttt{conv\_proj} for ViT-B/16, \texttt{features.0.0} for ConvNeXt-T, and \texttt{features.0.0} for EffNet-B0.
The selected perturbation probabilities are $p=0.02$ for ResNet-50, $p=0.2$ for RepVGG-A2, $p=0.15$ for ViT-B/16, $p=0.5$ for ConvNeXt-T, and $p=0.05$ for EffNet-B0.
The corresponding global relative distances are $1.31\%$, $1.72\%$, $1.99\%$, $1.56\%$, and $1.81\%$, respectively.
Accuracy drop is computed as the difference between the original and perturbed top-1 accuracy, and disagreement is computed as the percentage of samples whose top-1 prediction changes after perturbation.
All experiments use seed $42$.

\subsection{Copyright-infringing SpongeBob images}\label[Appendix]{app:rep-spongebob}

The diffusion U-Net of CommonCanvas-XL-C \cite{gokaslan2024commoncanvas} was trained \emph{ab initio} on a curated corpus of Creative Commons-licensed photographs and contains no SpongeBob copyrighted imagery. Despite this, the deployed model still reconstructs the character's signature visual features -- yellow skin, two front teeth, white shirt with a neck-tie. The leakage is plausibly mediated by generalization from training images that, although common licensed, contain some features of the copyrighted characters. Images are generated at $512{\times}512$ in \texttt{float16} with DPMSolverMultistep ($35$ steps). Specific seeds and guidance values are: \cref{fig:spongebob-b}, seed $42$, classifier-free guidance (CfG)  $15.0$; \cref{fig:spongebob-c}, seed $2$, CfG $10.0$. The negative prompt in both cases is \textit{``blurry, low quality, distorted, malformed, ugly''}.

\subsection{Harry Potter prompts}\label[Appendix]{app:rep-hp}

The reported prompt-response pairs in \cref{tab:hp-leak} are obtained from the publicly released \texttt{microsoft/Llama2-7b-WhoIsHarryPotter} model \cite{eldan2023whos}, an approximately-unlearned variant of \texttt{Llama-2-7b-chat-hf} from which knowledge of the \emph{Harry Potter} novels is claimed to have been ``erased''. Despite this claim, oblique probes that never name the series or its protagonist still elicit \emph{Harry Potter}-specific terminology and proper nouns. In the first pair, the model correctly recovers the rules of Quidditch and the term \emph{Quaffle} but misattributes the sport to Pratchett's \emph{Discworld}; in the second, asked to identify the missing Marauder of the Marauder's Map, the model volunteers the name of a \emph{Harry Potter}-specific institution (Beauxbatons Academy) instead of declining. Both responses are produced by greedy decoding with seed $42$ and are therefore reproducible.

\begin{table}[!htbp]
\centering
\caption{Two prompt--response pairs obtained from the publicly released \texttt{microsoft/Llama2-7b-WhoIsHarryPotter} model \cite{eldan2023whos}. \emph{Bold} marks tokens that betray retained knowledge of the \emph{Harry Potter} universe.}
\label{tab:hp-leak}
\setlength{\tabcolsep}{6pt}
\renewcommand{\arraystretch}{1.25}
\footnotesize
\begin{tabular}{p{0.38\linewidth} p{0.50\linewidth}}
\toprule
\textbf{Prompt} & \textbf{Model output} \\
\midrule
\textit{In a popular fantasy series, what is the magical sport played on broomsticks, with three hoops on each end and a small golden winged ball?}
&
The answer is: \textbf{Quaffle}.\newline\newline
\textbf{Quaffle} is a fictional sport that appears in several books of the ``\textbf{Discworld}'' series by Terry Pratchett. It is a game played on a \textbf{broomstick}-like object called a ``quaffleboard'', with three hoops on each end and a small golden winged ball. The objective of the game is to score points by hitting the ball through the opponent's hoops. \\
\midrule
\textit{Fill in the blank: ``The Marauder's Map was created by James Potter, Sirius Black, Remus Lupin, and \rule{1cm}{0.4pt}.''} &
Hint: This person is a student at \textbf{Beauxbatons Academy}. \\
\bottomrule
\end{tabular}
\end{table}

\section{Additional Experiments}\label[Appendix]{app:extra-results}

\subsection{Extra results: noise perturbation}\label[Appendix]{app:extra-result-gaussian-noise}

In \cref{tab:perturbation-noise}, we show the effect of Gaussian noise randomly added to all weights of the model. Each model is evaluated on a fixed test split: $5{,}000$ ImageNet validation images for all models. For each model, the noise standard deviation $\sigma$ is swept over $\{5{\times}10^{-4},\,10^{-3},\,2{\times}10^{-3},\,5{\times}10^{-3},\,10^{-2},\,2{\times}10^{-2},\,5{\times}10^{-2},\,10^{-1}\}$, and we report the smallest $\sigma$ that induces a decrease in accuracy of at least ten percentage points: ResNet-50, $\sigma{=}10^{-3}$; RepVGG-A2, $\sigma{=}2{\times}10^{-3}$; ViT-B/16, $\sigma{=}10^{-2}$; ConvNeXt-T, $\sigma{=}10^{-2}$; EffNet-B0, $\sigma{=}2{\times}10^{-2}$. All experiments use seed $42$.

\vspace{0.3cm}

\begin{table}[!htbp]
\centering
\caption{Top-1 accuracy (\%) of five pretrained models, before and after adding
i.i.d.\ Gaussian noise $\mathcal{N}(0,\sigma^{2})$; $\sigma \le 0.02$.}
\label{tab:perturbation-noise}
\footnotesize
\begin{tabular}{lccccc}
\toprule
& ResNet-50 & RepVGG-A2 & ViT-B/16 & ConvNeXt-T & EffNet-B0 \\
\midrule
Original (\%) & 81.20 & 76.96 & 81.48 & 82.74 & 77.80 \\
Noisy (\%)    & 31.68 & 0.08 & 41.20 & 40.70 & 19.68 \\
\bottomrule
\end{tabular}
\end{table}

\subsection{Extra generative examples}\label[Appendix]{app:copy-right-extra-results}

\begin{figure}[!htbp]
  \centering
  \begin{subfigure}[b]{0.34\linewidth}
    \centering
    \includegraphics[width=1.0\linewidth]{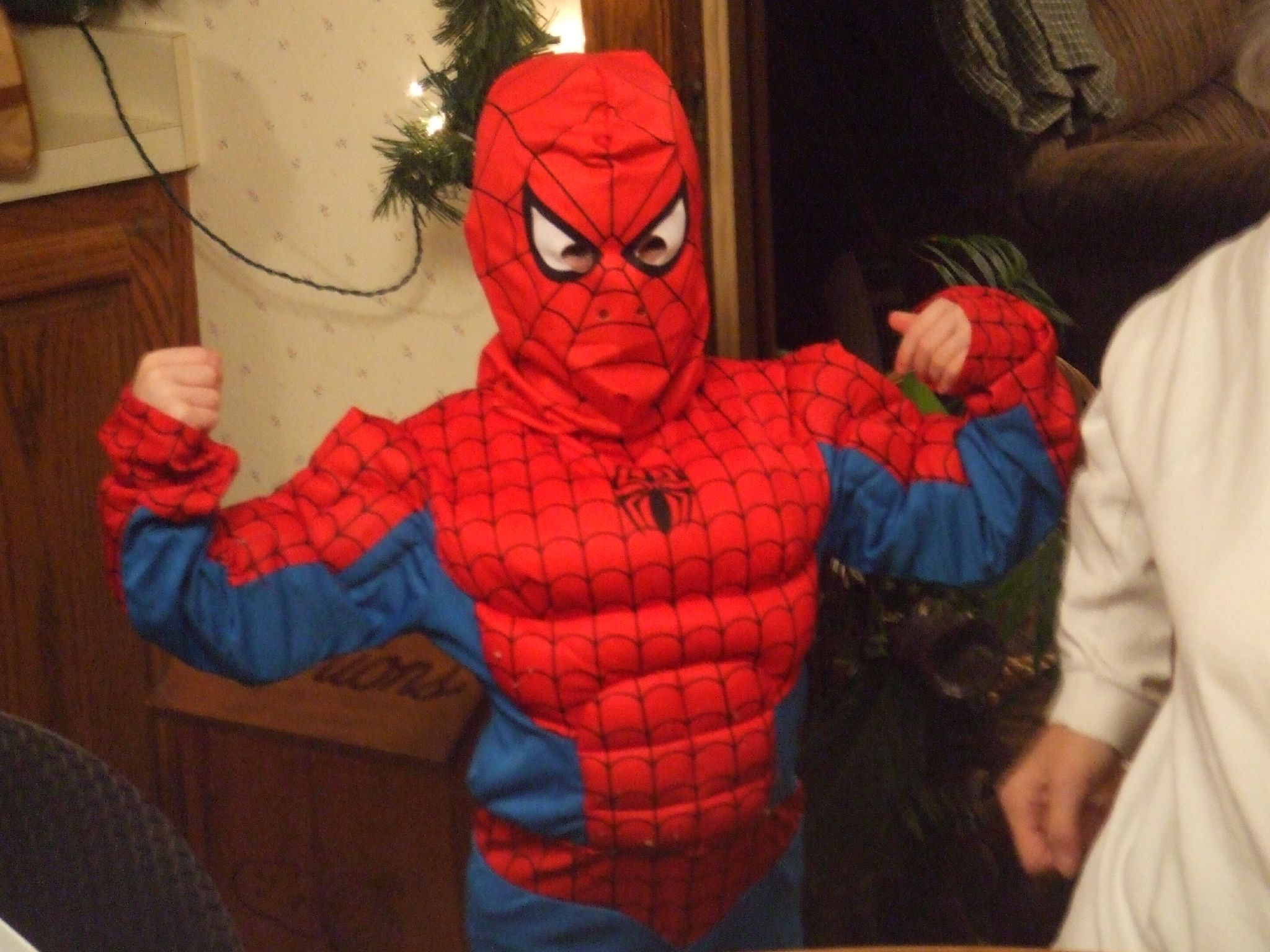}
    \caption{Spider-Man dataset instance.}
    \label{fig:spiderman-a}
  \end{subfigure}
  \hspace{0.5cm}
  \begin{subfigure}[b]{0.27\linewidth}
    \centering
    \includegraphics[width=0.95\linewidth]{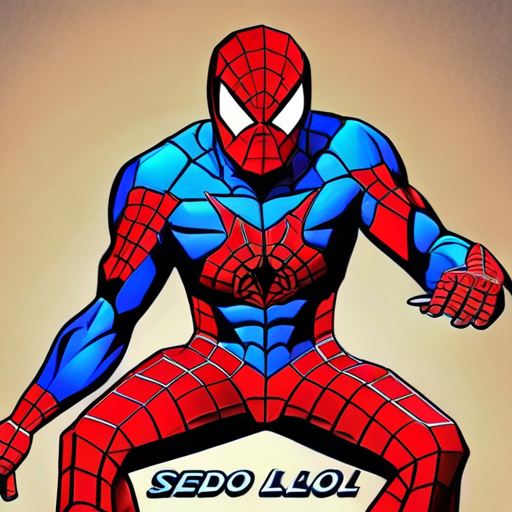}
    \caption{Prompt: \textit{``Spider Man solo''}.}
    \label{fig:spiderman-b}
  \end{subfigure}
  \hspace{0.5cm}
  \begin{subfigure}[b]{0.27\linewidth}
    \centering
    \includegraphics[width=0.95\linewidth]{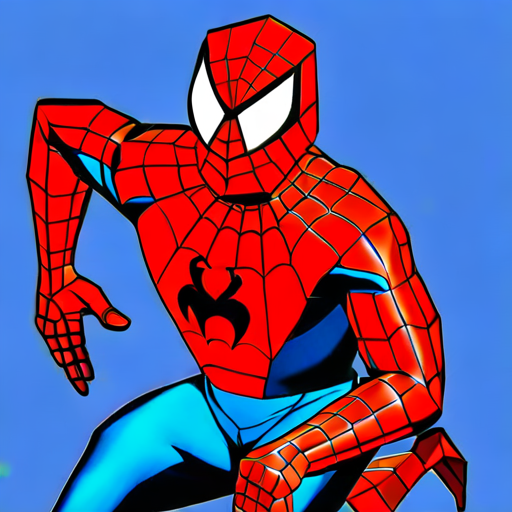}
    \caption{Prompt: \textit{``Spider Man''}.}
    \label{fig:spiderman-c}
  \end{subfigure}
  \caption{Spider-Man imagery generated by CommonCanvas-XL-C.}
  \label{fig:generative-copyright-spiderman}
\end{figure}

\begin{figure}[!htbp]
  \centering
  \begin{subfigure}[b]{0.27\linewidth}
    \centering
    \includegraphics[width=0.95\linewidth]{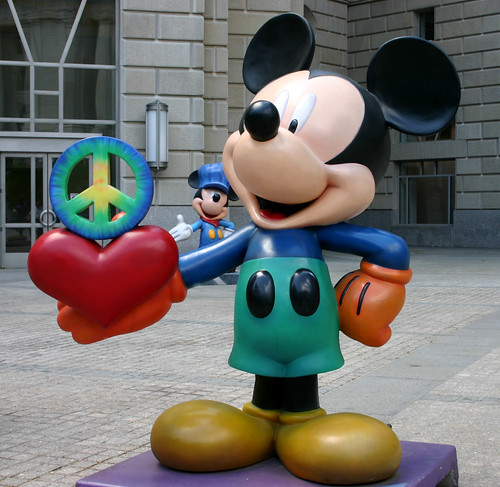}
    \caption{Mickey Mouse datasaet instance.}
    \label{fig:mickey-a}
  \end{subfigure}
  \hspace{0.5cm}
  \begin{subfigure}[b]{0.27\linewidth}
    \centering
    \includegraphics[width=0.95\linewidth]{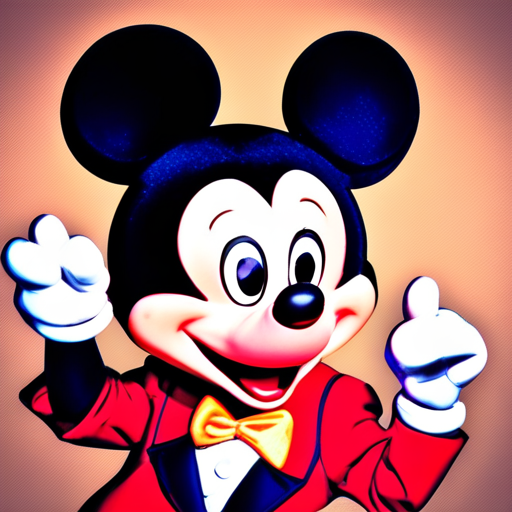}
    \caption{Prompt: \textit{``Mickey Mouse''}.\phantom{aaaa}}
    \label{fig:mickey-b}
  \end{subfigure}
  \hspace{0.5cm}
  \begin{subfigure}[b]{0.27\linewidth}
    \centering
    \includegraphics[width=0.95\linewidth]{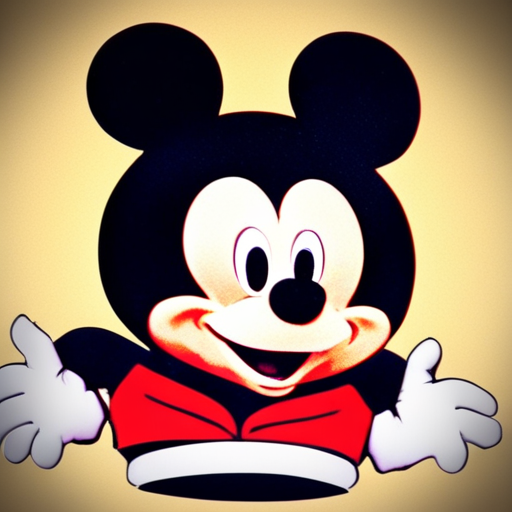}
    \caption{Prompt: \textit{``Mickey Mouse alone''}.}
    \label{fig:mickey-c}
  \end{subfigure}
  \caption{Mickey Mouse imagery generated by CommonCanvas-XL-C.}
  \label{fig:generative-copyright-mickey}
\end{figure}

\begin{figure}[!htbp]
  \centering
  \begin{subfigure}[b]{0.27\linewidth}
    \centering
    \includegraphics[width=0.72\linewidth]{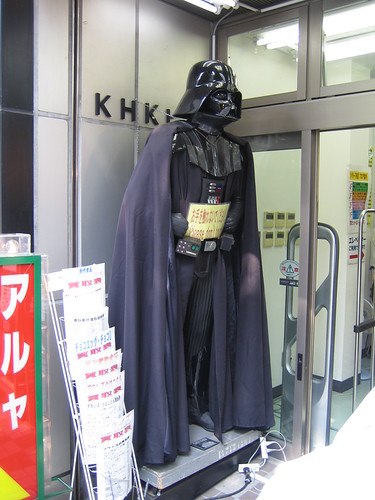}
    \caption{Darth Vader dataset instance. \phantom{, full body, dramatic backlight, photorealistic} \phantom{-----------------------}}
    \label{fig:darthvader-a}
  \end{subfigure}
  \hspace{0.5cm}
  \begin{subfigure}[b]{0.27\linewidth}
    \centering
    \includegraphics[width=0.95\linewidth]{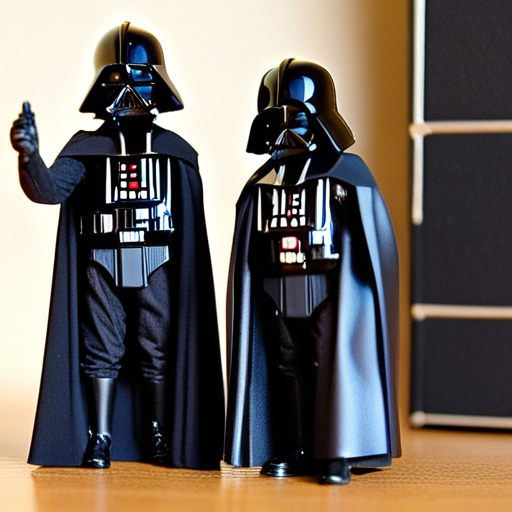}
    \caption{Prompt: \textit{``Darth Vader''}.\phantom{, full body, dramatic backlight, photorealistic} \phantom{-------------------}}
    \label{fig:darthvader-b}
  \end{subfigure}
  \hspace{0.5cm}
  \begin{subfigure}[b]{0.27\linewidth}
    \centering
    \includegraphics[width=0.95\linewidth]{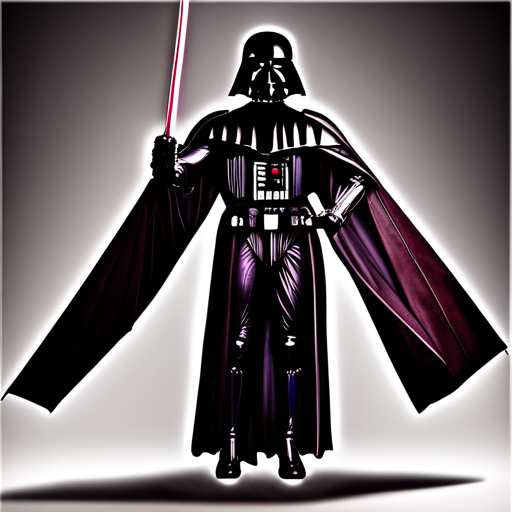}
    \caption{Prompt: \textit{``Darth Vader, full body, dramatic backlight, photorealistic''}.}
    \label{fig:darthvader-c}
  \end{subfigure}
  \caption{Darth Vader imagery generated by CommonCanvas-XL-C.}
  \label{fig:generative-copyright-darthvader}
\end{figure}

\begin{figure}[!htbp]
  \centering
  \begin{subfigure}[b]{0.205\linewidth}
    \centering
    \includegraphics[width=1.0\linewidth]{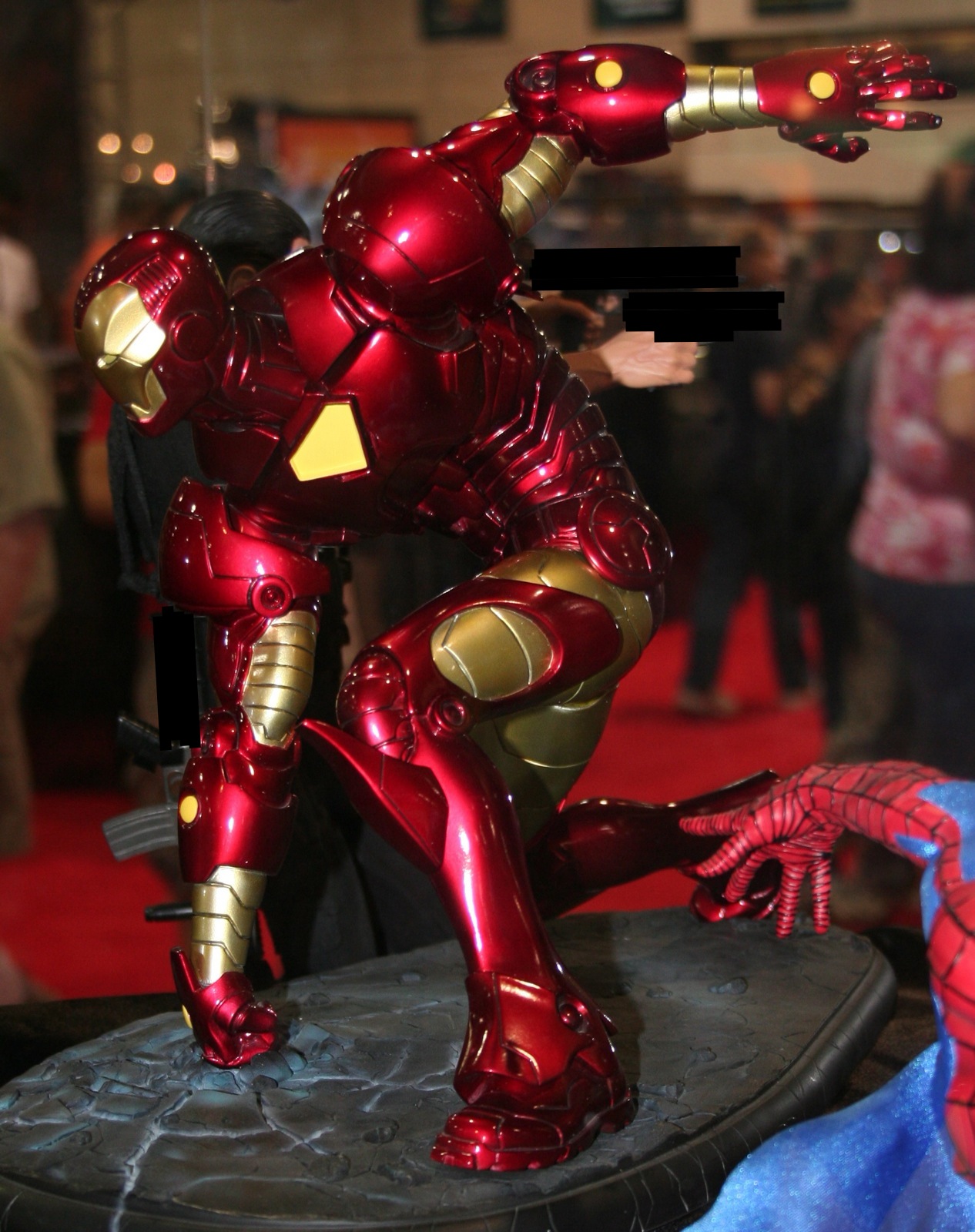}
    \caption{Iron Man dataset instance (censor bar).}
    \label{fig:ironman-a}
  \end{subfigure}
  \hspace{0.5cm}
  \begin{subfigure}[b]{0.27\linewidth}
    \centering
    \includegraphics[width=0.95\linewidth]{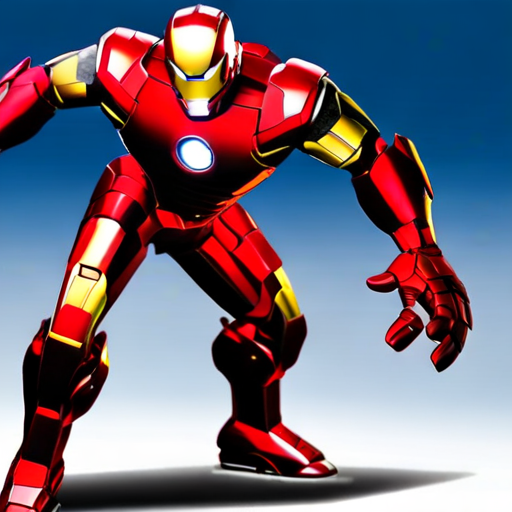}
    \caption{Prompt: \textit{``Iron Man''}. \phantom{-------------}}
    \label{fig:ironman-b}
  \end{subfigure}
  \hspace{0.5cm}
  \begin{subfigure}[b]{0.27\linewidth}
    \centering
    \includegraphics[width=0.95\linewidth]{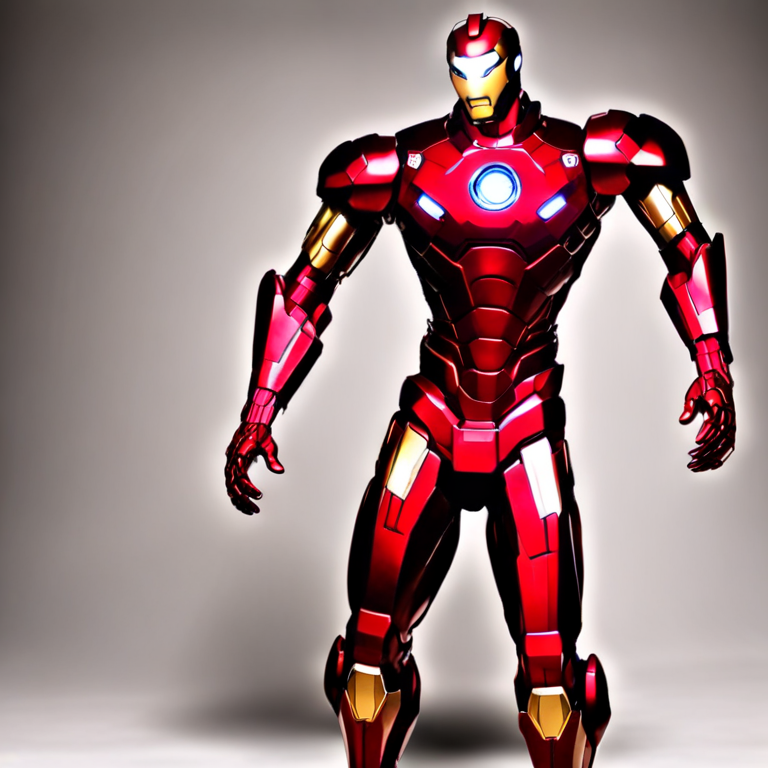}
    \caption{Prompt: \textit{``Iron-Man full body''}. \phantom{-------------}}
    \label{fig:ironman-c}
  \end{subfigure}
  \caption{Iron Man imagery generated by CommonCanvas-XL-C.}
  \label{fig:generative-copyright-ironman}
\end{figure}

In this appendix we present additional examples of copyright-infringing images generated by CommonCanvas-XL-C \cite{gokaslan2024commoncanvas}, whose \emph{image training corpus consists exclusively of Creative Commons-licensed photographs} (the CommonCatalog dataset) and contains no copyrighted imagery of Marvel, Disney, or Lucasfilm characters. Despite this, the model produces recognisable depictions of Spider-Man (\cref{fig:generative-copyright-spiderman}), Mickey Mouse (\cref{fig:generative-copyright-mickey}), Darth Vader (\cref{fig:generative-copyright-darthvader}), and Iron Man (\cref{fig:generative-copyright-ironman}). All images are generated with DPMSolverMultistep ($35$ steps) at resolution $512{\times}512$, except \cref{fig:ironman-c} which uses $768{\times}768$. The seeds and gradient-guidance scales are: \cref{fig:spiderman-b}, seed $123$, guidance $15.0$; \cref{fig:spiderman-c}, seed $15$, guidance $20.0$; \cref{fig:mickey-b,fig:mickey-c}, seed $123$, guidance $10.0$; \cref{fig:darthvader-b}, seed $42$, guidance $10.0$; \cref{fig:darthvader-c}, seed $1$, guidance $10.0$; \cref{fig:ironman-b}, seed $1$, guidance $15.0$; and \cref{fig:ironman-c}, seed $5$, guidance $17.5$. The negative prompt across all generations is \textit{``blurry, low quality, distorted, malformed, ugly''}.

\vspace{1cm}

\vspace{1cm}

\end{document}